\documentclass[]{article}
\usepackage{lmodern}
\usepackage{amssymb,amsmath}
\usepackage{ifxetex,ifluatex}
\usepackage{fixltx2e} %
\ifnum 0\ifxetex 1\fi\ifluatex 1\fi=0 %
  \usepackage[T1]{fontenc}
  \usepackage[utf8]{inputenc}
\else %
  \ifxetex
    \usepackage{mathspec}
  \else
    \usepackage{fontspec}
  \fi
  \defaultfontfeatures{Ligatures=TeX,Scale=MatchLowercase}
\fi
\IfFileExists{upquote.sty}{\usepackage{upquote}}{}
\IfFileExists{microtype.sty}{%
\usepackage{microtype}
\UseMicrotypeSet[protrusion]{basicmath} %
}{}
\usepackage[margin=1in]{geometry}
\usepackage{hyperref}
\hypersetup{unicode=true,
            pdftitle={Comparing Different Deep Learning Architectures for Classification of Chest Radiographs},
            pdfborder={0 0 0},
            breaklinks=true}
\usepackage{biblatex}

\usepackage{longtable,booktabs}
\usepackage{graphicx,grffile}
\makeatletter
\def\maxwidth{\ifdim\Gin@nat@width>\linewidth\linewidth\else\Gin@nat@width\fi}
\def\maxheight{\ifdim\Gin@nat@height>\textheight\textheight\else\Gin@nat@height\fi}
\makeatother
\setkeys{Gin}{width=\maxwidth,height=\maxheight,keepaspectratio}
\IfFileExists{parskip.sty}{%
\usepackage{parskip}
}{%
\setlength{\parindent}{0pt}
\setlength{\parskip}{6pt plus 2pt minus 1pt}
}
\ifx\paragraph\undefined\else
\let\oldparagraph\paragraph
\renewcommand{\paragraph}[1]{\oldparagraph{#1}\mbox{}}
\fi
\ifx\subparagraph\undefined\else
\let\oldsubparagraph\subparagraph
\renewcommand{\subparagraph}[1]{\oldsubparagraph{#1}\mbox{}}
\fi

\let\rmarkdownfootnote\footnote%
\def\footnote{\protect\rmarkdownfootnote}

\usepackage{titling}

  \title{Comparing Different Deep Learning Architectures for Classification of
Chest Radiographs}
    \pretitle{\vspace{\droptitle}\centering\huge}
  \posttitle{\par}
    \author{\newline \small \emph{Keno K. Bressem, Lisa Adams, Christoph Erxleben,
Bernd Hamm, Stefan Niehues, Janis Vahldiek} \bigskip  
\newline \emph{Department of Radiology, Charité Universitätsmedizin
Berlin}\\
\bigskip     \small\emph{\texttt{keno-kyrill.bressem(at)charite.de}}}
    \preauthor{\centering\large\emph}
  \postauthor{\par}
    \date{}
    \predate{}\postdate{}

\begin{document}
\maketitle
\begin{abstract}
Chest radiographs are among the most frequently acquired images in
radiology and are often the subject of computer vision research.
However, most of the models used to classify chest radiographs are
derived from openly available deep neural networks, trained on large
image-datasets. These datasets routinely differ from chest radiographs
in that they are mostly color images and contain several possible image
classes, while radiographs are greyscale images and often only contain
fewer image classes. Therefore, very deep neural networks, which can
represent more complex relationships in image-features, might not be
required for the comparatively simpler task of classifying grayscale
chest radiographs. We compared fifteen different architectures of
artificial neural networks regarding training-time and performance on
the openly available CheXpert dataset to identify the most suitable
models for deep learning tasks on chest radiographs. We could show, that
smaller networks such as ResNet-34, AlexNet or VGG-16 have the potential
to classify chest radiographs as precisely as deeper neural networks
such as DenseNet-201 or ResNet-151, while being less computationally
demanding.\\
\end{abstract}

\twocolumn

\hypertarget{introduction}{%
\section{Introduction}\label{introduction}}

Chest radiographs are among the most frequently used imaging procedures
in radiology. They have been widely employed in the field of computer
vision, as chest radiographs are a standardized technique and, if
compared to other radiological examinations such as computed tomography
or magnetic resonance imaging, contain a smaller group of relevant
pathologies. Although many artificial neural networks for the
classification of chest radiographs have been developed, it is still the
subject of intensive research. Only a few groups design their own
networks from scratch, but rather use already established architectures,
such as ResNet-50 or DenseNet-121 (with 50 and 121 representing the
number of layers within the respective neural network)
\autocite{resnet}\autocite{densenet}\autocite{irvin2019chexpert}\autocite{bustos2019}\autocite{rajpurkar2017}\autocite{pham2019}.
These neural networks have often been trained on large, openly available
datasets, such as ImageNet, and are therefore already able to recognize
numerous image features. When training a model for a new task, such as
the classification of chest radiographs, the use of pre-trained networks
may improve the training speed and accuracy of the new model, since
important image features that have already been learned can be
transferred to the new task and do not have to be learned again.
However, the feature space of freely available data sets such as
ImageNet differs from chest radiographs as they contain color images and
more categories. The ImageNet Challenge includes 1000 possible
categories per image, while CheXpert, a large freely available data set
of chest radiographs, only distinguishes between 14 categories (or
classes)\autocite{raghu2019}. Although the ImageNet challenge showed a
trend towards higher accuracies for deeper networks, this may not be
fully transferrable to radiology. In radiology, sometimes only limited
features of an image can be decisive for the diagnosis. Therefore,
images cannot be scaled down as much as desired, as the required
information would otherwise be lost. But, the more complex a neural
network architecture is, the more resources are required for training
and deployment of such an algorithm. As up-scaling the input-images
resolution exponentially increases memory usage during training for
large neural networks, that evaluate many parameters, the size of a mini
batch needs to be reduced earlier and more strongly, potentially
affecting optimizers such as stochastic gradient descent. Therefore, it
is currently not clear, which of the available artificial neural
networks designed for and trained on the ImageNet dataset will perform
the best for the classification of chest radiographs. The hypothesis of
this work is, that shallow networks are already sufficient for the
classification of radiographs and might even outperform deeper networks
while requiring lesser resources. Therefore, we systematically examine
the performance of fifteen openly available artificial neural network
architectures in order to identify the most suitable ones for the basic
classification of chest radiographs.

\hypertarget{methods}{%
\section{Methods}\label{methods}}

\hypertarget{data-preparation}{%
\subsection{Data preparation}\label{data-preparation}}

The free available CheXpert dataset consists of 224,316 chest
radiographs from 65,240 patients. Fourteen findings have been annotated
for each image: enlarged cardiomediastinum, cardiomegaly, lung opacity,
lung lesion, edema, consolidation, pneumonia, atelectasis, pneumothorax,
pleural effusion, pleural other, fracture and support devices. Hereby
the findings can be annotated as present (1), absent (NA) or uncertain
(-1). Similar to previous work on the classification of the CheXpert
dataset \autocite{irvin2019chexpert}\autocite{sabottke2020}, we trained
these networks on a subset of labels: cardiomegaly, edema,
consolidation, atelectasis and pleural effusion. As we only aim at
network comparison and not on maximal precision of a neural network, for
this analysis, each image with an uncertainty label was excluded, other
approaches such as zero imputation or self-training were also not
adopted. Furthermore, only frontal radiographs were used, leaving
135,494 images from 53,388 patients for training. CheXpert offers
additional dataset with 235 images (201 images after excluding
uncertainty labels and lateral radiographs), annotated by two
independent radiologists, which is intended as an evaluations dataset
and was therefore used for this purpose.

\hypertarget{data-augmentation}{%
\subsection{Data augmentation}\label{data-augmentation}}

For the first and second training session, the images were scaled to 320
x 320 pixels, using bilinear interpolation, and pixel values were
normalized. During training, multiple image-transformations were
applied: flipping of the images alongside the horizontal and vertical
axis, rotation of up to 10°, zooming of up to 110\%, adding of random
lightning or symmetric wrapping.

\hypertarget{model-training}{%
\subsection{Model training}\label{model-training}}

14 different convolutional neural networks (CNN) of five different
architectures (ResNet, DenseNet, VGG, SqueezeNet and AlexNet) were
trained on the CheXpert dataset
\autocite{resnet}\autocite{densenet}\autocite{vgg}\autocite{squeezenet}\autocite{alexnet}.
All training was done using the Python programming language
(\url{https://www.python.org}, version 3.8) with the PyTorch
(\url{https://pytorch.org}) and FastAI (\url{https://fast.ai}) libraries
on a workstation running on Ubuntu 18.04 with two Nvidia GeForce RTX
2080ti graphic cards (11 GB of RAM
each)\autocite{fastai}\autocite{pytorch}. In the first training session,
batch size was held constant at 16 for all models, while it was
increased to 32 for all networks in the second session. In the first two
sessions, each model was trained for eight epochs, whereas during the
first five epochs only the classification-head of each network was
trained. Thereafter, the model was unfrozen and trained as whole for
three additional epochs. Before training and after the first five
epochs, the optimal learning rate was determined \autocite{smith2017},
which was between 1e-1 and 1e-2 for the first five epochs and between
1e-5 and 1e-6 for the rest of the training. We trained one multilabel
classification head for each model. Since the performance of a neural
network can be subject to minor random fluctuations, the training was
repeated for a total of five times. The predictions on the validation
data set were then exported as comma separated values (CSV) for
evaluation.

\hypertarget{evaluation}{%
\subsection{Evaluation}\label{evaluation}}

Evaluation was performed using the ``R'' statistical environment
including the ``tidyverse'' and ``ROCR'' libraries
\autocite{R-base}\autocite{R-tidyverse}\autocite{rocr}.Predictions on
the validation dataset of the five models for each network architecture
were pooled so that the models could be evaluated as a consortium. For
each individual prediction as well as the pooled predictions, receiver
operation characteristic (ROC) curves and precision recall curves (PRC)
were plotted and the areas under each curve were calculated (AUROC and
AUPRC). AUROC and AUPRC were chosen as they enable a comparison of
different models, independent of a chosen threshold for the
classification.

\hypertarget{results}{%
\section{Results}\label{results}}

The CheXpert validation dataset consists out of 234 studies of 200
patients, not used for training with no uncertainty-labels. After
excluding lateral radiographs (n = 32), 202 images of 200 patients
remained. The dataset presents class imbalances (\% positives for each
finding: cardiomegaly 33\%, edema 21\%, consolidation 16\%, atelectasis
37\%, pleural effusion 32\%), so that the AUPRC as well as the AUROC can
be considered equally important measurements for the performance of the
network. The performance of the tested networks is compared to the AUROC
reported by Irvin et al.\autocite{irvin2019chexpert}.However, only
values for AUROC, but not for AUPRC, are provided there. In most cases,
the best results were achieved with a batch size of 32, so all the
information provided below refers to models trained with this batch
size. Results achieved with smaller batch sizes of 16 will be explicitly
mentioned.

\hypertarget{area-under-the-receiver-operating-characteristic-curve}{%
\subsection{Area under the Receiver Operating Characteristic
Curve}\label{area-under-the-receiver-operating-characteristic-curve}}

Deeper artificial neural networks generally achieved higher AUROC values
than shallow networks (Table 1 and Figures 1-3). Regarding the pooled
AUROC for the detection of the five pathologies, ResNet-152 (0.882),
DenseNet-161 (0.881) and ResNet-50 (0.881) performed best (Irvin et
al.~CheXpert baseline 0.889)\autocite{irvin2019chexpert}. Broken down
for individual findings, the most accurate detection of atelectasis was
achieved by ResNet-18 (0.816, batch size 16), ResNet-101 (0.813, batch
size 16), VGG-19 (0.813, batch size 16) and ResNet-50 (0.811). For
detection of cardiomegaly, the best four models surpassed the CheXpert
baseline of 0.828 (ResNet-34 0.840, ResNet-152 0.836, DenseNet-161
0.834, ResNet-50 0.832). For congestion, the highest AUROC was achieved
using ResNet-152 (0.917), ResNet-50 (0.916) and DenseNet-161 (0.913).
Pulmonary edema was most accurately detected using DenseNet-161 (0.923),
DenseNet-169 (0.922) and DenseNet-201 (0.922). For pleural effusion, the
four best models were ResNet-152 (0.937), ResNet-101 (0.936), ResNet-50
(0.934) and DenseNet-169 (0.934), all of which performed superior to the
CheXpert baseline of 0.928.

\hypertarget{area-under-the-precision-recall-curve}{%
\subsection{Area under the Precision Recall
Curve}\label{area-under-the-precision-recall-curve}}

For AUPRC, shallower artificial neural networks could achieve higher
values than deeper network-architectures (Table 2 and Figures 4-6). The
highest pooled values for the AUPRC were achieved by training VGG-16
(0.709), AlexNet (0.701) and ResNet-34 (0.688). For atelectasis, CGG-16
and AlexNet both achieved the highest AUPRC of 0.732, followed by
Resnet-35 with 0.652. Cardiomegaly was most accurately detected by
SqueezeNet 1.0 (0.565), Alexnet-152 (0.565) and Vgg-13 (0.563).
SqueezNet 1.0 also achieved the highest AUPRC values for consolidation
(0.815) followed by ResNet-152 (0.810) and ResNet-50 (0.809). The best
classifications of pulmonary edema were achieved by DenseNet-169,
DenseNet-161 (both 0.743) and DenseNet-201 (0.742). Finally, for pleural
effusion ResNet-101 and ResNet-152 achieved the highest AUPRC of 0.591,
followed by ResNet-50 (0.590).

\hypertarget{overall-best-performance}{%
\subsection{Overall best Performance}\label{overall-best-performance}}

Considering both AUROC and AUPRC, the best performance was achieved by
VGG-16 (AUROC: 0.856, AUPRC: 0.709), ResNet-34 (AUROC: 0.872, AUPRC:
0.688) and AlexNet (AUROC: 0.839, AUPRC: 0.701), all with a batch size
of 32.

\hypertarget{training-time}{%
\subsection{Training time}\label{training-time}}

Fourteen different network-architectures were trained 10 times each with
a multilabel-classification head (five times each for batch size of 16
or 32 and an input-image resolution of 320 x 320 pixels) and once with a
binary classification head for each finding, resulting in 210 individual
training runs. Overall, training took 340 hours. As to be expected, the
training of deeper networks required more time than the training of
shallower networks. For an image resolution of 320 x 320 pixels, the
training of AlexNet required the least amount of time with a time per
epoch of 2:29 to 2:50 minutes and a total duration of 20 minutes for
a batch size of 32. Using a smaller batch size of 16, the time per epoch
raised to 2:59 - 3:06 minutes and a total duration of 24 minutes. In
contrast, using a batch size of 16, training of a DenseNet-201 took the
longest with 5:11 hours and epochs requiring 41 minutes. For a batch size of 32, training a DenseNet-169 required the
largest amount of time with 3:06 hours (epochs between 21 and
27 minutes). Increasing the batch size from 16 to 32 lead to an
average acceleration of training by 29.9\% \(\pm\) 9.34\%. Table 3 gives
an overview of training times.

\hypertarget{discussion}{%
\section{Discussion}\label{discussion}}

In the present work, different architectures of artificial neural
networks are analyzed with respect to their performance for the
classification of chest radiographs. We could show that more complex
neural networks do not necessarily perform better than shallow networks.
Instead, an accurate classification of chest radiographs may be achieved
with comparably shallow networks, such as AlexNet (8 layers), ResNet-34
or VGG-16, which surpass even complex deep networks such as ResNet-150
or DenseNet-201.\\
The use of smaller neural networks has the advantage, that hardware
requirements and training time are lower compared to deeper networks.
Shorter training times allow to test more hyperparameters, simplifying
the overall training process. Lower hardware requirements also enable
the use of increased image resolutions. This could be of relevance for
the evaluation of chest radiographs with a generic resolution of 2048 x
2048 px to 4280 x 4280 px, where specific findings, such as small
pneumothorax, require larger resolutions of input-images, because
otherwise the crucial information regarding their presence could be lost
due to a downscaling. Furthermore, shorter training times might simplify
the integration of improvement methods into the training data, such as
the implementation of `human in the loop' annotations. `Human in the
loop' implies that the training of a network is supervised by a human
expert, who may intervene and correct the network at critical steps. For
example, the human expert can check the misclassifications with the
highest loss for incorrect labels, thus effectively reducing label
noise. With shorter training times, such feedback loops can be executed
faster. In the CheXpert dataset, which was used as a groundwork for the
present analysis, labels for the images were generated using a
specifically developed natural language processing tool, which did not
produce perfect labels. For example, the F1 scores for the mention and
subsequent negation of cardiomegaly were 0.973 and 0.909, and the F1
score for an uncertainty label was 0.727. Therefore, it can be assumed,
that there is a certain amount of noise in the training data, which
might affect the accuracy of the models trained on it. Implementing a
human-in-the loop approach for partially correcting the label noise
could further improve performance of networks trained on the CheXpert
dataset \autocite{karimi2019deep}. Our findings differ from applied
techniques used in previous literature, where deeper network
architectures, mainly a DenseNet-121, were used instead of small
networks to classify the CheXpert data set
\autocite{pham2019}\autocite{allaouzi2019}\autocite{sabottke2019}. The
authors of the CheXpert dataset achieved an average overall AUROC of
0.889 \autocite{irvin2019chexpert}, using a DenseNet-121 which was not
surpassed by any of the models used in our analysis, although
differences between the best performing networks and the CheXpert
baseline were smaller than 0.01.. It should be noted, however, that in
our analysis the hyperparameters for the models were probably not
selected as precise as in the original CheXpert paper by Irvin et al.,
since the focus of this work was more on comparing the architectures and
not on the complete optimization of one specific network. Still, we
identified model, which achieved higher AUROC values in two of the five
findings (cardiomegaly and effusion). Pham et al.~also used a
DenseNet-121 as the basis for their model and proposed the most accurate
model of the CheXpert dataset with a mean AUROC of 0.940 for the five
selected findings \autocite{pham2019}. The good results are probably due
to the hierarchical structure of the classification framework, which
takes into account correlations between different labels, and the
application of a label-smoothing technique, which also allows the use of
uncertainty labels (which were excluded in our present work). Allaouzi
et al.~similarly used a DenseNet-121 and created three different models
for the classification of the CheXpert and ChestX-ray14, yielding an AUC
of 0.72 for atelectasis, 0.87-0.88 for cardiomegaly, 0.74-0.77 for
consolidation, 0.86-0.87 for edema and 0.90 for effusion
\autocite{allaouzi2019}. Except for cardiomegaly, we achieved better
values with several models (e.g.~ResNet-34, ResNet-50, AlexNet, VGG-16).
We would interpret this as proof that complex deep networks are not
necessarily superior to more shallow networks for chest x-ray
classification. At least for the CheXpert dataset it seems that methods
optimizing the handling of uncertainty labels and hierarchical
structures of the data are important to improve model performance.
Sabottke et al.~trained a ResNet-32 for classification of chest
radiographs and therefore are one of the few groups using a smaller
network \autocite{sabottke2019}. With an AUROC of 0.809 for atelectasis,
0.925 for cardiomegaly, 0.888 for edema and 0.859 for effusion, their
network performed not as good as some of our tested networks. Raghu et
al.~employed a ResNet-50, an Inception-v3 as well as a custom-designed
small network. Similar to our findings, they observed, that smaller
networks showed a comparable performance to deeper networks
\autocite{raghu2019}.

\hypertarget{conclusion}{%
\subsubsection{Conclusion}\label{conclusion}}

In the present work, we could show that smaller artificial neural
networks for the classification of chest radiographs can perform
similar, or even surpass deeper and very deep neural networks. In
contrast to many previous studies, which mostly used a DenseNet-121, we
achieved the best results with up to 95\% smaller networks. Using
smaller networks therefore has the advantage that that they have lower
hardware requirements, as they require less GPU RAM and can be trained
faster without loss of performance.

\onecolumn

\hypertarget{tables}{%
\section{Tables}\label{tables}}

\hypertarget{table-1-area-under-the-receiver-operating-characteristic-curve}{%
\subsection{Table 1 Area under the Receiver Operating Characteristic
Curve}\label{table-1-area-under-the-receiver-operating-characteristic-curve}}

\begin{longtable}[]{@{}lrrrrrrr@{}}
\toprule
Network & Batchsize & Atelectasis & Cardiomegaly & Consolidation & Edema
& Effusion & Pooled\tabularnewline
\midrule
\endhead
\textbf{CheXpert baseline} & \textbf{16} & \textbf{0.818} & \textbf{0.828} & \textbf{0.938} & \textbf{0.934} & \textbf{0.928} & \textbf{0.889}\tabularnewline
ResNet-18 & 16 & 0.816 & 0.797 & 0.905 & 0.868 & 0.899 & 0.857\tabularnewline
ResNet-34 & 16 & 0.799 & 0.798 & 0.902 & 0.891 & 0.905 & 0.859\tabularnewline
ResNet-50 & 16 & 0.798 & 0.799 & 0.890 & 0.880 & 0.913 & 0.856\tabularnewline
ResNet-101 & 16 & 0.813 & 0.810 & 0.905 & 0.889 & 0.907 & 0.865\tabularnewline
ResNet-152 & 16 & 0.801 & 0.809 & 0.908 & 0.896 & 0.916 & 0.866\tabularnewline
DenseNet-121 & 16 & 0.809 & 0.794 & 0.895 & 0.883 & 0.906 & 0.857\tabularnewline
DenseNet-161 & 16 & 0.800 & 0.817 & 0.885 & 0.900 & 0.923 & 0.865\tabularnewline
DenseNet-169 & 16 & 0.805 & 0.795 & 0.898 & 0.891 & 0.909 & 0.860\tabularnewline
DenseNet-201 & 16 & 0.805 & 0.812 & 0.891 & 0.886 & 0.916 & 0.862\tabularnewline
AlexNet & 16 & 0.790 & 0.755 & 0.857 & 0.894 & 0.881 & 0.835\tabularnewline
SqueezeNet-1.0 & 16 & 0.761 & 0.755 & 0.833 & 0.907 & 0.885 & 0.828\tabularnewline
SqueezeNet-1.1 & 16 & 0.767 & 0.764 & 0.880 & 0.903 & 0.879 & 0.839\tabularnewline
VGG-13 & 16 & 0.798 & 0.752 & 0.886 & 0.867 & 0.872 & 0.835\tabularnewline
VGG-16 & 16 & 0.809 & 0.766 & 0.892 & 0.879 & 0.883 & 0.846\tabularnewline
VGG-19 & 16 & 0.811 & 0.786 & 0.901 & 0.890 & 0.884 & 0.854\tabularnewline
ResNet-18 & 32 & 0.796 & 0.822 & 0.908 & 0.903 & 0.911 & 0.868\tabularnewline
ResNet-34 & 32 & 0.797 & \textbf{0.840} & 0.903 & 0.902 & 0.919 & 0.872\tabularnewline
ResNet-50 & 32 & 0.811 & \textbf{0.832} & 0.916 & 0.913 & \textbf{0.934} & 0.881\tabularnewline
ResNet-101 & 32 & 0.797 & 0.823 & 0.911 & 0.915 & \textbf{0.936} & 0.876\tabularnewline
ResNet-152 & 32 & 0.802 & \textbf{0.836} & 0.917 & 0.920 & \textbf{0.937} & 0.882\tabularnewline
DenseNet-121 & 32 & 0.808 & 0.828 & 0.879 & 0.904 & 0.926 & 0.869\tabularnewline
DenseNet-161 & 32 & 0.809 & \textbf{0.834} & 0.913 & 0.923 & 0.928 & 0.881\tabularnewline
DenseNet-169 & 32 & 0.809 & 0.816 & 0.900 & 0.922 & \textbf{0.934} & 0.876\tabularnewline
DenseNet-201 & 32 & 0.795 & 0.820 & 0.904 & 0.922 & 0.931 & 0.874\tabularnewline
AlexNet & 32 & 0.791 & 0.768 & 0.856 & 0.894 & 0.886 & 0.839\tabularnewline
SqueezeNet-1.0 & 32 & 0.773 & 0.769 & 0.880 & 0.913 & 0.895 & 0.846\tabularnewline
SqueezeNet-1.1 & 32 & 0.785 & 0.789 & 0.895 & 0.904 & 0.898 & 0.854\tabularnewline
VGG-13 & 32 & 0.800 & 0.762 & 0.883 & 0.896 & 0.907 & 0.850\tabularnewline
VGG-16 & 32 & 0.798 & 0.776 & 0.890 & 0.911 & 0.906 & 0.856\tabularnewline
VGG-19 & 32 & 0.787 & 0.790 & 0.879 & 0.911 & 0.916 & 0.857\tabularnewline
\bottomrule
\end{longtable}

\textbf{Table 1} shows the different areas under the receiver operating
characteristic curve (AUROC) for each of the network architectures and
individual finding as well as the pooled AUROC per model. According to
the pooled AUROC, ResNet-152, ResNet-50 und DenseNet-161 were the best
models, while SqueezeNet and AlexNet showed the poorest performance. For
cardiomegaly, ResNet-34, ResNet-50, ResNet-152 and DenseNet-161 could
surpass the CheXpert baseline provided by Irvin et al.~ResnEt-50,
ResNet-101, ResNet-152 and DenseNet-169 could also surpass the CheXpert
baseline for pleural effusion. A batch size of 32 often lead to better
results compared to a batch size of 16.

\newpage

\hypertarget{table-2-area-under-the-precision-recall-curve}{%
\subsection{Table 2 Area under the Precision Recall
Curve}\label{table-2-area-under-the-precision-recall-curve}}

\begin{longtable}[]{@{}lrrrrrrr@{}}
\toprule
Network & Batchsize & Atelectasis & Cardiomegaly & Consolidation & Edema
& Effusion & Pooled\tabularnewline
\midrule
\endhead
ResNet-18 & 16 & 0.500 & 0.559 & 0.806 & 0.727 & 0.580 &
0.634\tabularnewline
ResNet-34 & 16 & 0.506 & 0.560 & 0.804 & 0.735 & 0.580 &
0.637\tabularnewline
ResNet-50 & 16 & 0.501 & 0.557 & 0.802 & 0.733 & 0.585 &
0.636\tabularnewline
ResNet-101 & 16 & 0.499 & 0.558 & 0.765 & 0.735 & 0.582 &
0.628\tabularnewline
ResNet-152 & 16 & 0.503 & 0.559 & 0.808 & 0.737 & 0.584 &
0.638\tabularnewline
DenseNet-121 & 16 & 0.503 & 0.554 & 0.802 & 0.733 & 0.580 &
0.634\tabularnewline
DenseNet-161 & 16 & 0.501 & 0.557 & 0.799 & 0.736 & 0.587 &
0.636\tabularnewline
DenseNet-169 & 16 & 0.500 & 0.560 & 0.805 & 0.733 & 0.582 &
0.636\tabularnewline
DenseNet-201 & 16 & 0.320 & 0.555 & 0.445 & 0.734 & 0.582 &
0.527\tabularnewline
AlexNet & 16 & 0.543 & 0.565 & 0.490 & 0.733 & 0.577 &
0.582\tabularnewline
SqueezeNet-1.0 & 16 & 0.509 & 0.565 & 0.425 & 0.736 & 0.576 &
0.562\tabularnewline
SqueezeNet-1.1 & 16 & 0.505 & 0.563 & 0.400 & 0.733 & 0.575 &
0.555\tabularnewline
VGG-13 & 16 & 0.502 & 0.563 & 0.761 & 0.726 & 0.574 &
0.625\tabularnewline
VGG-16 & 16 & 0.501 & 0.559 & 0.797 & 0.733 & 0.577 &
0.633\tabularnewline
VGG-19 & 16 & 0.500 & 0.558 & 0.808 & 0.731 & 0.577 &
0.635\tabularnewline
ResNet-18 & 32 & 0.502 & 0.557 & 0.805 & 0.736 & 0.582 &
0.636\tabularnewline
ResNet-34 & 32 & 0.652 & 0.556 & 0.806 & 0.737 & 0.585 &
0.667\tabularnewline
ResNet-50 & 32 & 0.497 & 0.555 & 0.809 & 0.740 & 0.590 &
0.638\tabularnewline
ResNet-101 & 32 & 0.500 & 0.558 & 0.808 & 0.740 & 0.591 &
0.639\tabularnewline
ResNet-152 & 32 & 0.502 & 0.559 & 0.810 & 0.741 & 0.591 &
0.641\tabularnewline
DenseNet-121 & 32 & 0.500 & 0.558 & 0.793 & 0.736 & 0.587 &
0.635\tabularnewline
DenseNet-161 & 32 & 0.499 & 0.556 & 0.808 & 0.743 & 0.589 &
0.639\tabularnewline
DenseNet-169 & 32 & 0.499 & 0.556 & 0.805 & 0.743 & 0.588 &
0.638\tabularnewline
DenseNet-201 & 32 & 0.502 & 0.555 & 0.808 & 0.742 & 0.589 &
0.639\tabularnewline
AlexNet & 32 & 0.720 & 0.562 & 0.789 & 0.731 & 0.578 &
0.676\tabularnewline
SqueezeNet-1.0 & 32 & 0.354 & 0.562 & 0.815 & 0.738 & 0.580 &
0.610\tabularnewline
SqueezeNet-1.1 & 32 & 0.506 & 0.563 & 0.804 & 0.731 & 0.577 &
0.636\tabularnewline
VGG-13 & 32 & 0.501 & 0.560 & 0.799 & 0.735 & 0.578 &
0.635\tabularnewline
VGG-16 & 32 & 0.732 & 0.561 & 0.804 & 0.739 & 0.582 &
0.684\tabularnewline
VGG-19 & 32 & 0.501 & 0.562 & 0.800 & 0.740 & 0.585 &
0.638\tabularnewline
\bottomrule
\end{longtable}

\textbf{Table 2} shows the area under the precision recall curve (AUPRC)
for all networks and findings. In contrast to the AUROC, where deeper
models achieved higher values, shallower networks yielded the best
results for AUPRC (ResNet-24, AlexNet, VGG-16). DenseNet-201 and
Squeezenet showed the lowest AUPRC values. Again, a batch size of 32
appeared to deliver better results compared to a batch size of 16.
\newpage

\hypertarget{table-3-duration-of-training}{%
\subsection{Table 3 Duration of
Training}\label{table-3-duration-of-training}}

\begin{longtable}[]{@{}lrll@{}}
\toprule
Network & Batchsize & Duration/Epoch & Duration/Training\tabularnewline
\midrule
\endhead
ResNet-18 & 16 & 6 min & 50 min\tabularnewline
ResNet-34 & 16 & 10 min & 1  h 13 min\tabularnewline
ResNet-50 & 16 & 11 min - 13 min & 1  h 40 min\tabularnewline
ResNet-101 & 16 & 19 min - 25 min & 2  h 47 min\tabularnewline
ResNet-152 & 16 & 27 min - 28 min & 4  h 7 min\tabularnewline
SqueezeNet-1.0 & 16 & 4 min - 6 min & 39 min\tabularnewline
SqueezeNet-1.1 & 16 & 4 min & 37 min\tabularnewline
AlexNet & 16 & 3 min & 24 min\tabularnewline
VGG-13 & 16 & 12 min & 1  h 49 min\tabularnewline
VGG-16 & 16 & 20 min - 21 min & 2  h 14 min\tabularnewline
VGG-19 & 16 & 24 min & 2  h 40 min\tabularnewline
DenseNet-121 & 16 & 23 min - 25 min & 3  h 7 min\tabularnewline
DenseNet-169 & 16 & 31 min - 34 min & 4  h 21 min\tabularnewline
DenseNet-161 & 16 & 29 min - 36 min & 4  h 17 min\tabularnewline
DenseNet-201 & 16 & 41 min & 5  h 11 min\tabularnewline
ResNet-18 & 32 & 4 min & 31 min\tabularnewline
ResNet-34 & 32 & 5 min - 7 min & 45 min\tabularnewline
ResNet-50 & 32 & 8 min & 1  h 16 min\tabularnewline
ResNet-101 & 32 & 13 min & 2  h 8 min\tabularnewline
ResNet-152 & 32 & 21 min - 26 min & 2  h 58 min\tabularnewline
SqueezeNet-1.0 & 32 & 3 min - 4 min & 28 min\tabularnewline
SqueezeNet-1.1 & 32 & 3 min & 25 min\tabularnewline
AlexNet & 32 & 2 min - 3 min & 20 min\tabularnewline
VGG-13 & 32 & 10 min - 14 min & 1  h 31 min\tabularnewline
VGG-16 & 32 & 17 min & 1  h 47 min\tabularnewline
VGG-19 & 32 & 13 min & 2  h 2 min\tabularnewline
DenseNet-121 & 32 & 12 min - 16 min & 1  h 49 min\tabularnewline
DenseNet-169 & 32 & 17 min & 2  h 25 min\tabularnewline
DenseNet-161 & 32 & 21 min - 27 min & 3  h 6 min\tabularnewline
DenseNet-201 & 32 & 20 min & 2  h 52 min\tabularnewline
\bottomrule
\end{longtable}

\textbf{Table 3} provides an overview of training time per epoch
(duration/epoch) and an overall training-time (duration/training) for
each neural network. The times given are the average of five training
runs rounded to the nearest minute.

\newpage

\hypertarget{figures}{%
\section{Figures}\label{figures}}

\hypertarget{receiver-operating-characteristic-curves}{%
\subsection{Receiver Operating Characteristic
Curves}\label{receiver-operating-characteristic-curves}}

Figures 1, 2 and 3 display the ROC-curves for all models. The colored
lines represent a single training, black lines represent the pooled
performance over five trainings.

\hypertarget{figure-1}{%
\subsubsection{Figure 1}\label{figure-1}}

\includegraphics{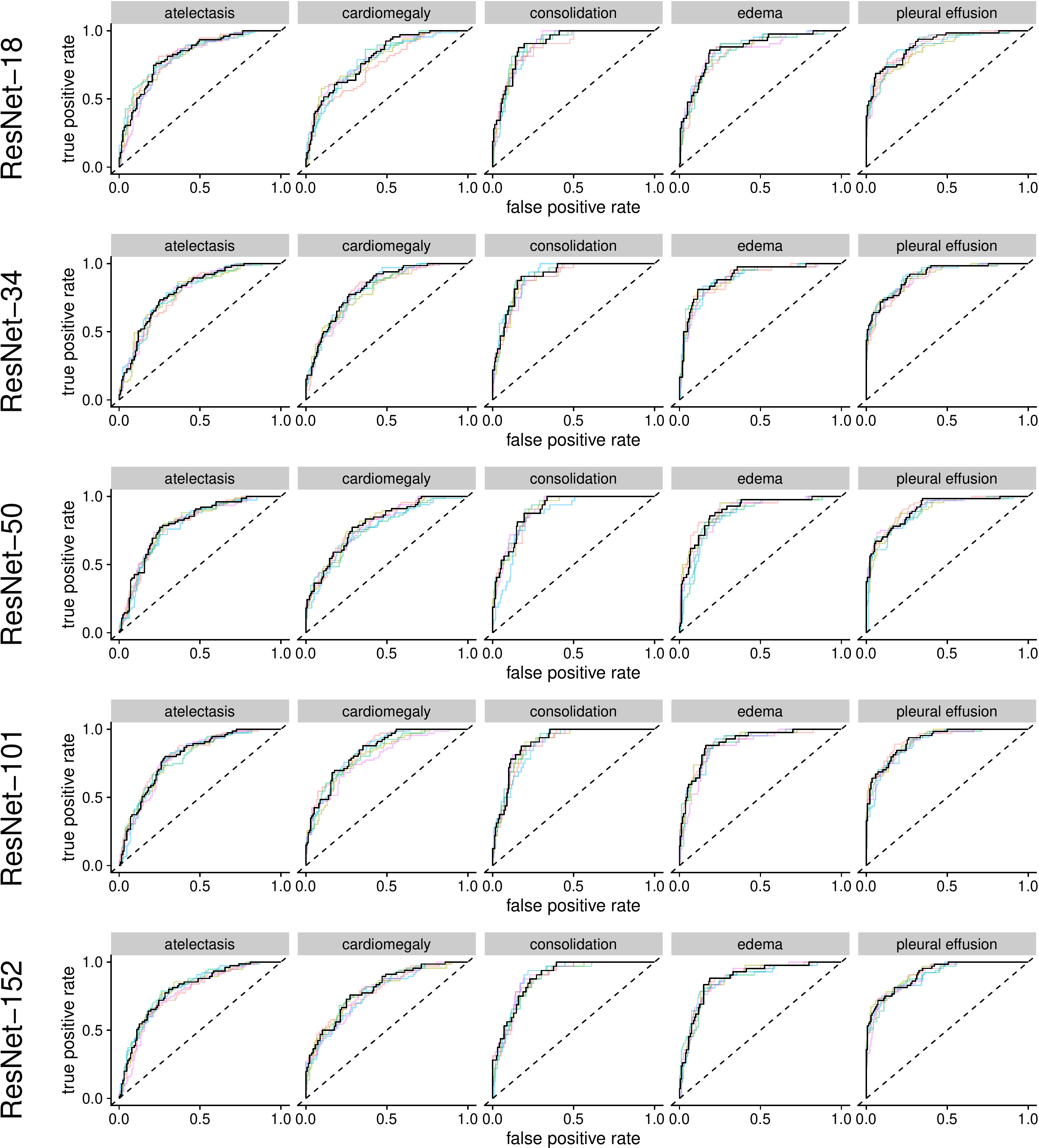}

\newpage

\hypertarget{figure-2}{%
\subsubsection{Figure 2}\label{figure-2}}

\includegraphics{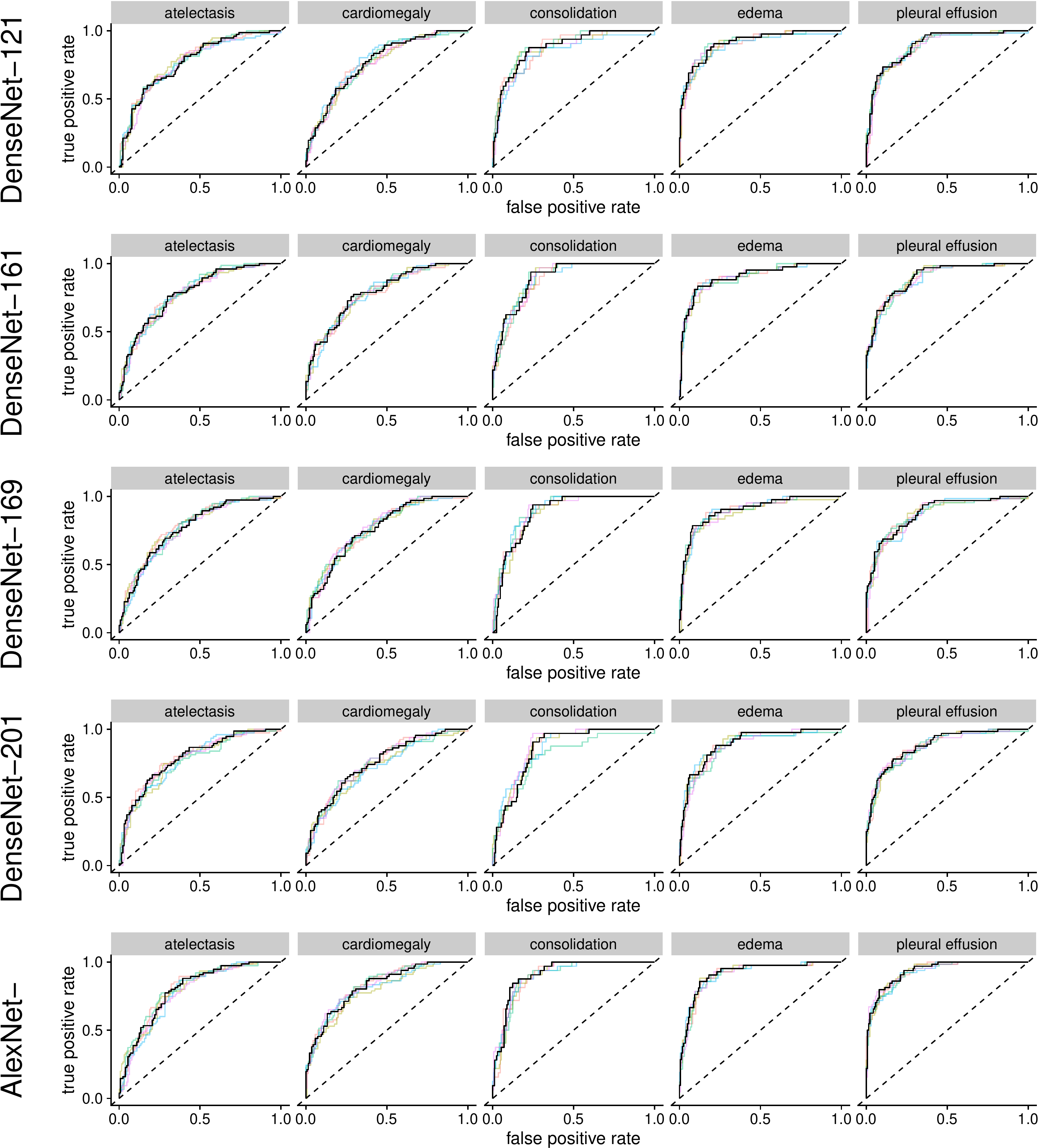}

\newpage

\hypertarget{figure-3}{%
\subsubsection{Figure 3}\label{figure-3}}

\includegraphics{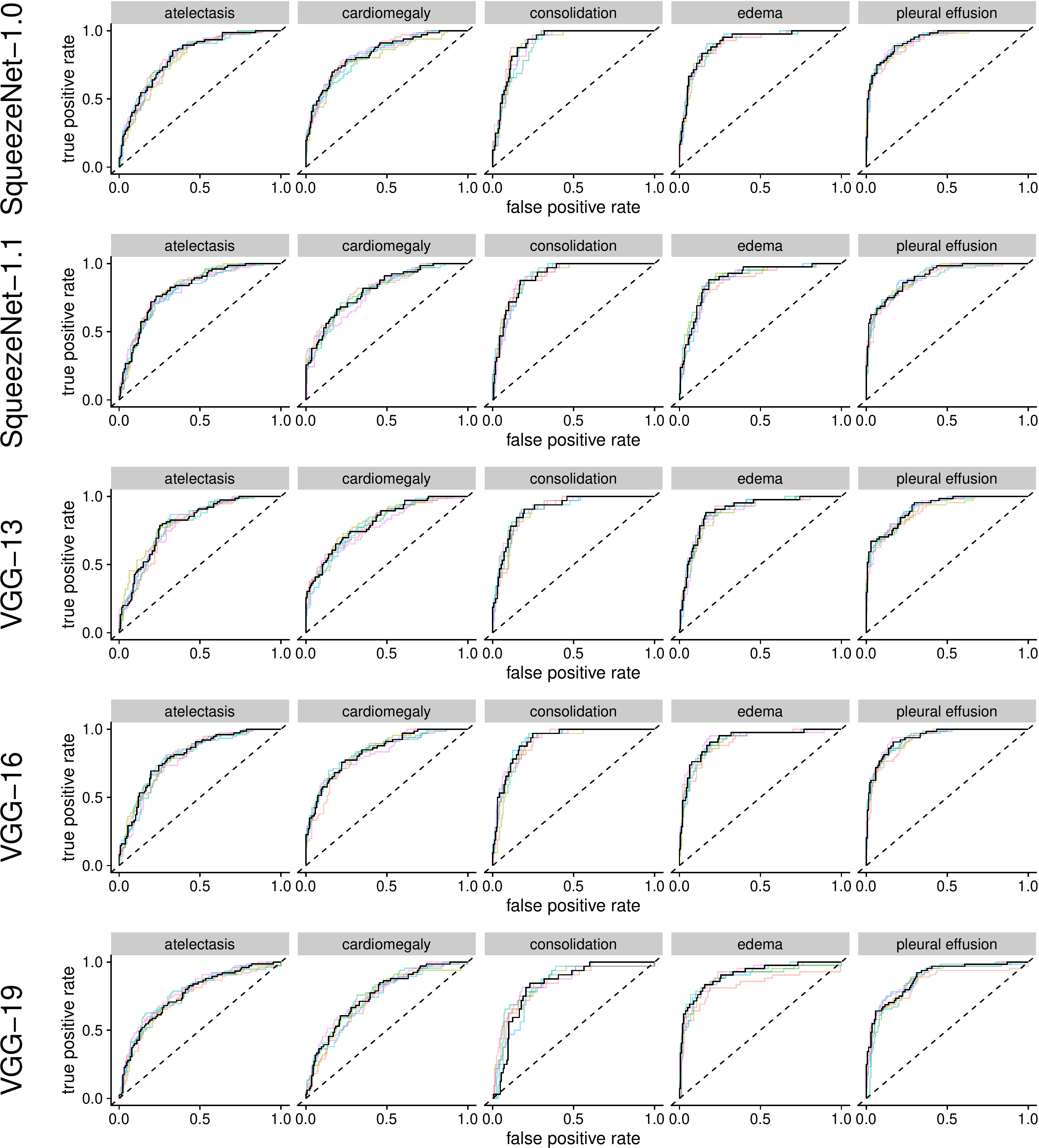}

\newpage

\hypertarget{precision-recall-curves}{%
\subsection{Precision Recall Curves}\label{precision-recall-curves}}

Figures 1, 2 and 3 display the precision recall curves for all models.
The colored lines represent a single training, black lines represent the
pooled performance over five trainings.

\hypertarget{figure-4}{%
\subsubsection{Figure 4}\label{figure-4}}

\includegraphics{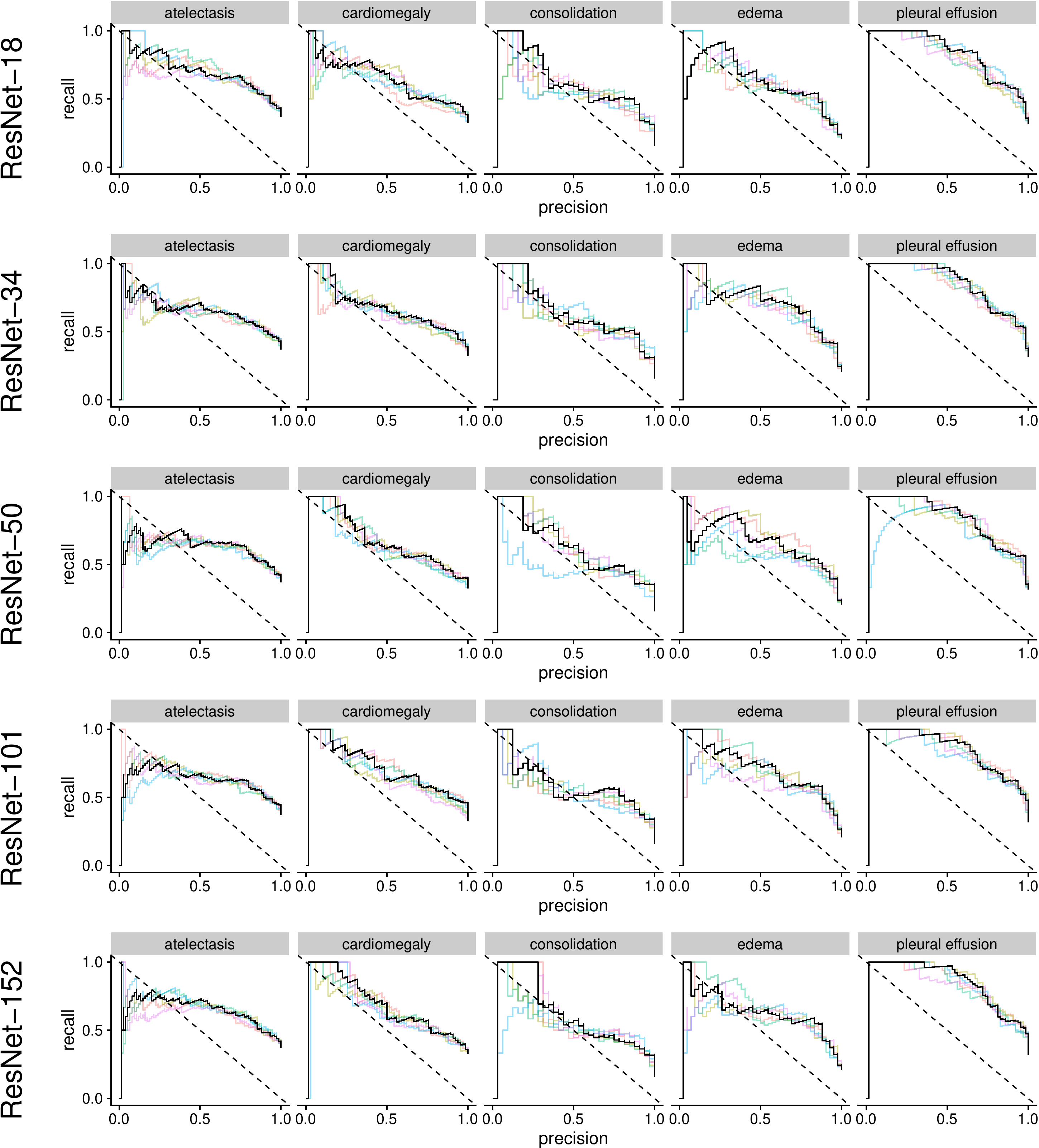}

\newpage

\hypertarget{figure-5}{%
\subsubsection{Figure 5}\label{figure-5}}

\includegraphics{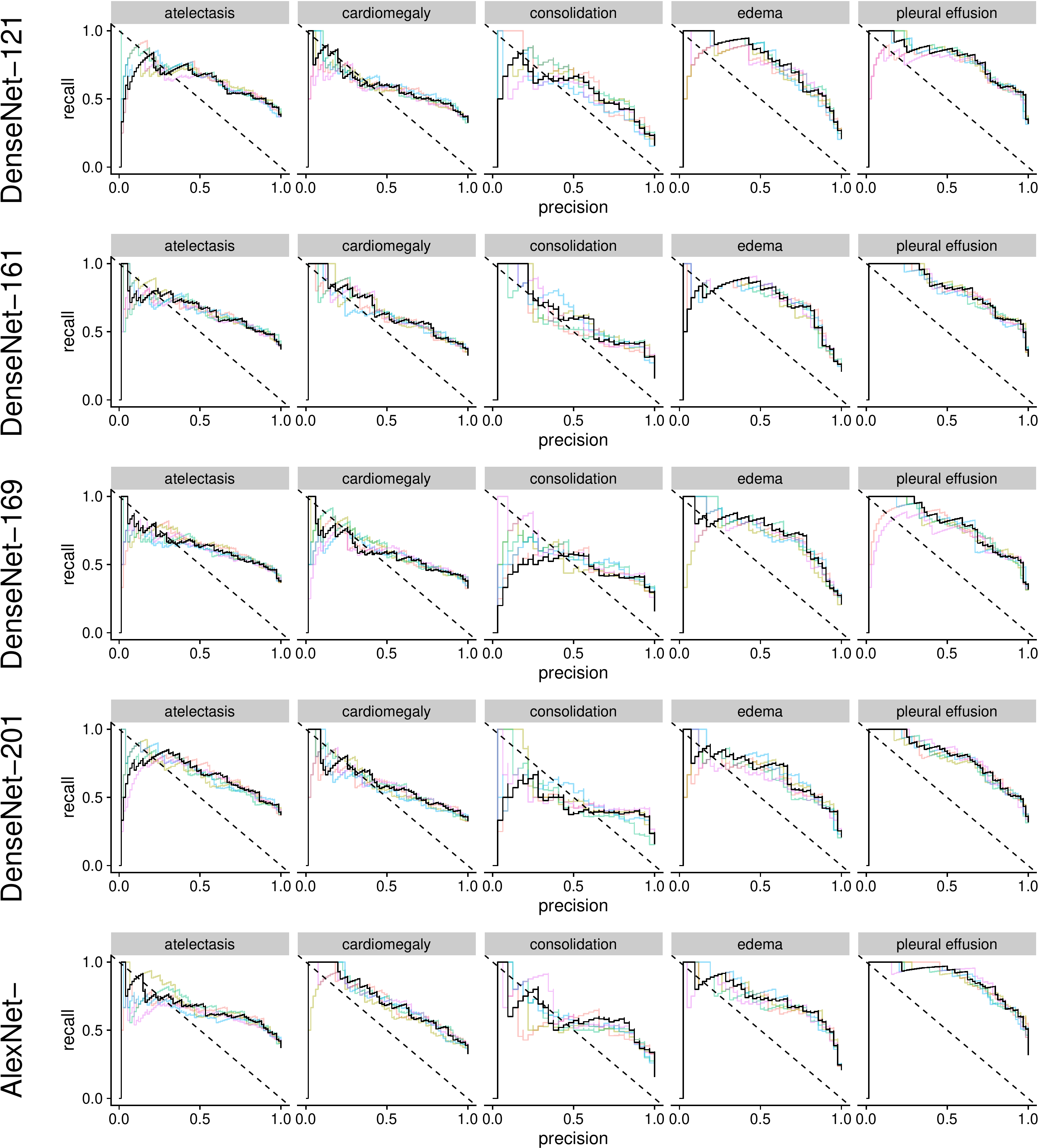}

\newpage

\hypertarget{figure-6}{%
\subsubsection{Figure 6}\label{figure-6}}

\includegraphics{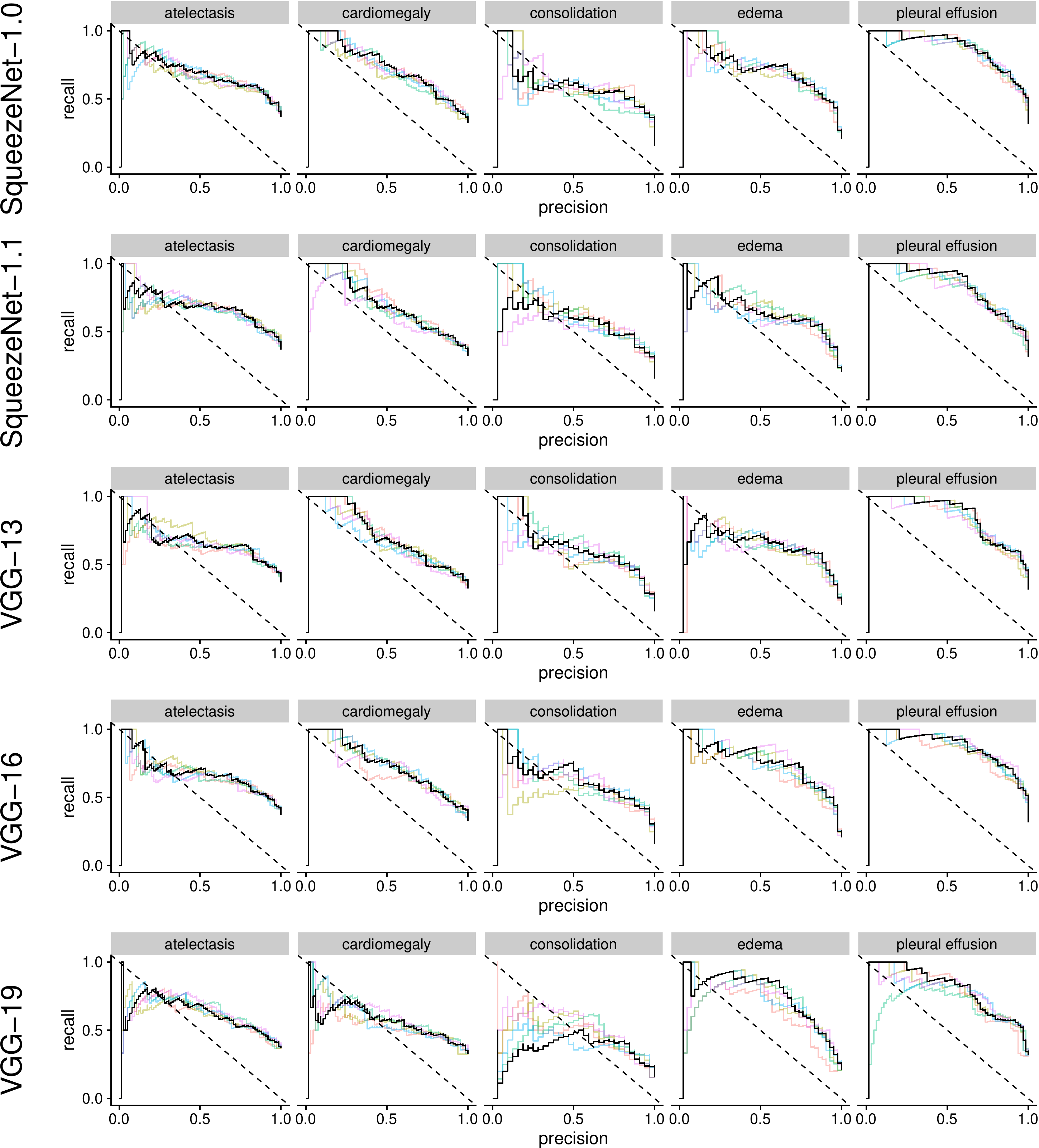}

\twocolumn

\printbibliography[title=References]

\end{document}